\documentclass[letterpaper, 10 pt, conference]{ieeeconf}

\IEEEoverridecommandlockouts   

\usepackage[utf8]{inputenc} 
\usepackage[T1]{fontenc}    
\usepackage{url}            
\usepackage{booktabs}       
\usepackage{amsfonts}       
\usepackage{graphicx}%
\usepackage{amsmath}
\usepackage{amssymb}
\usepackage{xcolor}
\usepackage{accents}
\usepackage{caption}
\usepackage{cite}
\usepackage{xcolor}
\usepackage{multirow}
\usepackage{siunitx}
\usepackage{amsmath,bm}
\usepackage{algorithm}
\usepackage{algpseudocode}
\usepackage{caption}
\usepackage{subcaption}
\usepackage{stfloats}
\usepackage{soul}
\setstcolor{red}
\usepackage[colorlinks=true, urlcolor=red, breaklinks=true]{hyperref}
\usepackage{breakurl}

\newtheorem{lemma}{Lemma}

\newtheorem{remark}{\textbf{Remark}}

\renewcommand{\baselinestretch}{.98}

\title{
Distributed and Consistent Multi-Robot Visual-Inertial-Ranging Odometry on Lie Groups
}
\author{%
    Ziwei Kang and Yizhi Zhou%
    \thanks{Z. Kang is affiliated with the School of Control and Computer Engineering, North China Electric Power University, Beijing, VA 102206, China. Y. Zhou is affiliated with the Department of Electrical and Computer Engineering, George Mason University, Fairfax, VA 22030, USA (e-mail: {\tt yzhou26@gmu.edu}). He is the corresponding author.}%
}

\begin{document}
\maketitle
\thispagestyle{empty}
\pagestyle{empty}

\begin{abstract}
Reliable localization is a fundamental requirement for multi-robot systems operating in GPS-denied environments. Visual–inertial odometry (VIO) provides lightweight and accurate motion estimation but suffers from cumulative drift in the absence of global references. Ultra-wideband (UWB) ranging offers complementary global observations, yet most existing UWB-aided VIO methods are designed for single-robot scenarios and rely on pre-calibrated anchors, which limits their robustness in practice. This paper proposes a distributed collaborative visual–inertial–ranging odometry (DC-VIRO) framework that tightly fuses VIO and UWB measurements across multiple robots. Anchor positions are explicitly included in the system state to address calibration uncertainty, while shared anchor observations are exploited through inter-robot communication to provide additional geometric constraints. By leveraging a right-invariant error formulation on Lie groups, the proposed approach preserves the observability properties of standard VIO, ensuring estimator consistency. Simulation results with multiple robots demonstrate that DC-VIRO significantly improves localization accuracy and robustness, while simultaneously enabling anchor self-calibration in distributed settings.
\end{abstract}

\section{INTRODUCTION}
Accurate positioning and mutual perception are essential for multi-robot systems, especially in GPS-denied indoor environments. Unlike single robots that rely solely on local sensors, multi-robot systems can improve accuracy by sharing relative measurements and joint observations of the environment \cite{Lee_2023}, \cite{Zhu_2021, MI2012}. Many approaches build on visual-inertial odometry (VIO) \cite{nisar2019vimo,9794476,li2019real}, which fuses camera and IMU data and can be extended with inter-robot information to further enhance accuracy. Although lightweight and precise, VIO suffers from cumulative drift due to the lack of global references, and this drift can propagate across robots through relative pose estimations. To address this issue, ultra-wideband (UWB) ranging provides a promising solution by introducing anchor-to-robot distance measurements as global observations \cite{yin2024uwb, WC2017}.

For UWB-aided VIO systems, the key to achieving accurate localization lies in how UWB information is fused with visual-inertial measurements \cite{ZW2021, NTUviral}. A well-designed UWB-aided VIO system should ensure not only error-bounded but also consistency when fusing ranging measurements \cite{KD2018}. Inconsistent estimators tend to underestimate system uncertainty, which in turn degrades overall localization accuracy \cite{huang2009first, HJ2014}. Early methods were typically loosely coupled, assuming that anchor positions had been precisely calibrated offline \cite{nguyen2021flexible, NTUviral}. However, such approaches often require manual calibration and neglect the uncertainty in anchor placement, which can lead to inconsistency and reduced robustness in practical deployments \cite{WC2017}. Subsequent studies have explored tightly coupled frameworks to better exploit the complementary information of UWB and VIO \cite{YB2021, JS2023}. Nevertheless, most of these methods were developed for single-robot scenarios and cannot be directly extended to multi-robot systems, particularly in distributed settings. This paper therefore focuses on the problem of consistently fusing UWB measurements within a tightly coupled framework for distributed multi-robot localization.

Unlike single-robot UWB-aided VIO, the multi-robot case introduces additional challenges and opportunities. When multiple robots simultaneously observe the same UWB anchor, the ranging measurements naturally form shared observations across robots \cite{CVIO}. Such shared anchors not only provide extra geometric constraints for improving localization accuracy but also enable the joint calibration of anchor positions, which are often uncertain in practice\cite{cao2021vir}. To address these challenges, this paper develops a fully distributed collaborative visual–inertial–ranging odometry (DC-VIRO) framework that fuses VIO and UWB information in a tightly coupled manner. Anchor positions are explicitly included in the system state to account for calibration uncertainty, while the right-invariant error formulation on Lie groups guarantees that the proposed system preserves the same four unobservable directions as the original VIO, ensuring observability consistency. The main contributions of this paper are:
\begin{itemize}
\item We propose a fully distributed DC-VIRO system that leverages inter-robot communication and shared UWB anchors to construct additional geometric constraints, thereby improving both collaborative localization accuracy and anchor calibration.
\item We define the system state on a Lie group and prove, using the right-invariant error properties, that DC-VIRO retains the same four unobservable directions as standard VIO, ensuring estimator consistency.
\item We validate the feasibility and effectiveness of the proposed system through extensive simulations experiments.
\end{itemize}

\noindent\textit{Notations:}
Let $\mathbb{R}$ denote the set of real numbers.
We use $\mathbf{I}_n$ and $\mathbf{0}_n$ to denote the $n \times n$ identity and zero matrices, respectively. For any matrix $\mathbf{A} \in \mathbb{R}^{m \times n}$, its transpose is denoted by $\mathbf{A}^\top \in \mathbb{R}^{n \times m}$. The notation $\text{diag}(\mathbf{A}_1, \mathbf{A}_2, \ldots, \mathbf{A}_k)$ represents a block-diagonal matrix with $\mathbf{A}_1, \mathbf{A}_2, \ldots, \mathbf{A}_k$ along the diagonal and zeros elsewhere.

\begin{figure}
  \centering
  \includegraphics[width=0.4\textwidth]{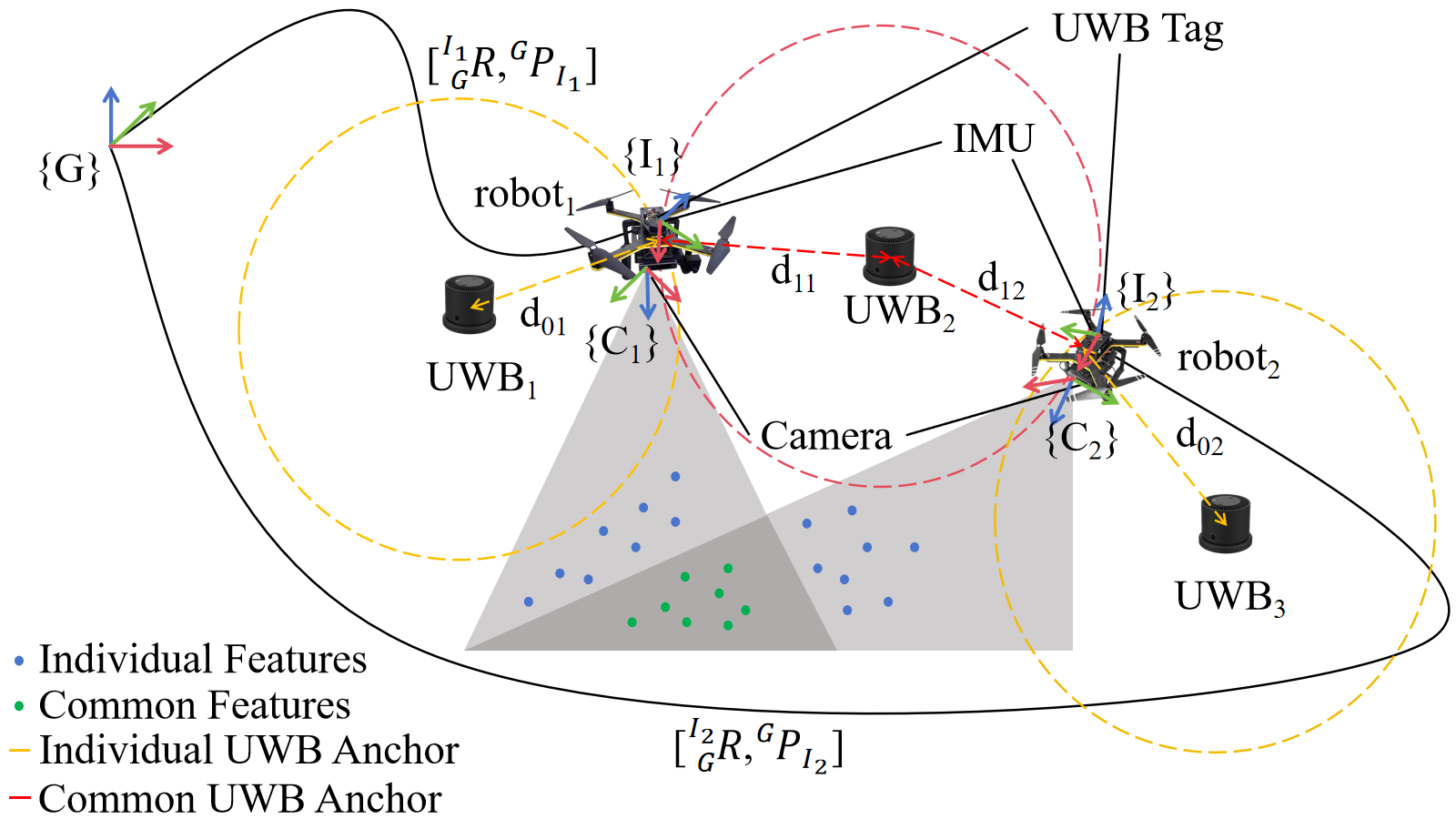}
  \caption{Cooperative visual-range-inertial navigation setup.}
  \label{Fig_1}
\end{figure}

\section{System Design}
In this section, we introduce the DC-VIRO system. It is a Multi-State Constraint Kalman Filter (MSCKF) framework formulated on matrix Lie groups for multi-robot cooperative localization. Specifically, we consider a system of $N$ robots as shown in Figure \ref{Fig_1}. Each robot is equipped with an IMU, UWB, and a camera. We first define the IMU and camera frames at time $t_k$ as $I_{i,k}$ and $C_{i,k}$, respectively, where $i$ represents the robot index, and $G$ denotes the global frame. Moreover, robots can communicate and share \textit{inter-robot information} with their neighbors to improve overall accuracy. As illustrated in Fig.~\ref{Fig_1}, two types of inter-robot information are considered in this work: UWB anchor observations and visual feature observations. For UWB anchors, each robot obtains ranging measurements between its UWB tag and surrounding anchors. When these anchors are observed only by a single robot, they are treated as individual anchor observations; when multiple robots observe the same anchor simultaneously, the ranging information becomes a common anchor observation that can be exploited for inter-robot alignment and joint estimation. Similarly, for visual features, a robot may detect and track features visible only to itself, which are regarded as individual feature observations, while features observed by multiple robots are treated as common feature observations and serve as shared constraints that improve the consistency and accuracy of cooperative localization.

The primary objective of the system is to simultaneously estimate the state of the robot's (IMU's) and the positions of the UWB anchors using the robot's own sensor measurements and the local information obtained through communication with neighboring robots. It is worth noting that, unlike the standard MSCKF framework formulated in vector space, the proposed DC-VIRO system operates directly on matrix Lie group representations. By leveraging the inherent manifold structure, the system enhances the consistency and accuracy of the estimator.

\subsection{System state}
Each robot $i$ at time $t_k$ has a state that includes the current IMU state, the positions of UWB anchors observed by itself, and sliding window information of $m$ historical IMU clones as:
\begin{align}\label{eq_state}
\mathbf{X}_{i,k}&=\left(
\mathbf X_{I_{i,k}}, \mathbf X_{C_{i}} \right)\nonumber\\
\mathbf X_{C_{i}}&=\left(\mathbf C_{i,k-1},  \cdots, \mathbf C_{i,k-m} \right)
\end{align}
Here, the IMU state and the UWB anchor state are jointly represented as $\mathbf{X}_{I_{i,k}}=(\mathbf T_{i,k}, \mathbf B_{i,k})\in SE_{2+L}(3)\times \mathbb R^6$, where $SE_{2+L}(3)$ represents the augmented Special Euclidean Group in three dimension, and $L$ represents the number of UWB anchors observed by the robot. For simplicity, we assume that the observation information of each robot contains only the information of one UWB anchor, $L=1$ in the state vector $X_{I_{i,k}}$. which can be easily generalizable to multi-anchor cases based on the definitions of $SE_{2+L}(3)$ group. Specifically, the state representation $\mathbf T_{i,k}\in SE_{2+L}(3)$ is defined as
\begin{align}
\mathbf T_{i,k}&=\left[
\begin{array}{c|c}
{^{G}_{I_{i,k}}\mathbf{R}} & 
^G\mathbf{v}_{I_{i,k}} \, ^G\mathbf{p}_{I_{i,k}} \, ^G\mathbf{p}_{u}\\
\hline
\mathbf 0_{3}&\mathbf I_{3}
\end{array}
\right]\in\ SE_3(3), 
\end{align}
that contains the IMU's rotation ${^G_{I_{i,k}}\mathbf{R}}\in SO(3)$ from the inertial frame ${I_{i,k}}$ to global frame ${G}$ at time $t_k$, and IMU's velocity ${^G{\mathbf v}_{I_{i,k}}}$ and position ${^G{\mathbf p}_{I_{i,k}}}$ in the global frame. It is important to highlight that, in addition to the IMU states, $\mathbf T_{i,k}$ also contains the UWB anchors' positions $^G \mathbf p_u$ since it is also unknown parameters that need to be estimated. 
$\mathbf B_{i,k}=\begin{bmatrix} \mathbf b_{\omega_{i,k}}^\top&\mathbf b_{a_{i,k}}^\top \end{bmatrix}^\top\in\mathbb R^6$ consists the IMU's gyroscope bias $\mathbf b_{\omega_{i,k}}$ and accelerometer bias 
$\mathbf b_{a_{i,k}}$.
For $l\in\{k-1, \cdots, k-m\}$, 
$\mathbf{C}_l\in SE(3)$,  denotes the historical clones of the IMU’s pose at time $t_l$ given by
\begin{align}
\mathbf{C}_l=\begin{bmatrix}
{^G_{I_l}\mathbf R} & ^G\mathbf p_{I_l}\\
\mathbf 0_{1\times 3} & 1
\end{bmatrix}    
\end{align}
By defining $\hat {\mathbf X}_{i,k}=\left(\hat {\mathbf X}_{I_{i,k}}, \hat{\mathbf X}_{C_{i}}\right)$ as the state estimate of $\mathbf X_{i,k}$, we formulate the error state $\widetilde{\mathbf X}_{i,k}$ as
\begin{align}
\widetilde{\mathbf X}_{i,k} &= \left(\hat{\mathbf T}_{i,k} \mathbf T_{i,k}^{-1}, \hat{\mathbf B}_{i,k}-\mathbf B_{i,k}, \hat{\mathbf C}_{i,l} \mathbf C_ {i,l}^{-1}\right)\nonumber\\
\hat{\mathbf C}_{i,l} \mathbf C_ {i,l}^{-1} &= \left(\hat{\mathbf C}_{i,k-1} \mathbf C_ {i,k-1}^{-1}, \cdots,\hat{\mathbf C}_{i,k-m} \mathbf C_{i,k-m}^{-1}\right)
\end{align}
where $\hat{\mathbf T}_{i,k} \mathbf T_{i,k}^{-1}\in SE_3(3)$ denotes the right invariant error given by
\begin{align}
\hat{\mathbf T}_{i,k} \mathbf T_{i,k}^{-1}&=
\left[
\begin{array}{c|c}
{^{G}_{I_{i,k}}{\mathbf {\widetilde R}}} & 
\bm\Gamma_{i,1} \, \bm\Gamma_{i,2} \, \bm\Gamma_{i,3}\\
\hline
\mathbf 0_{3}&\mathbf I_{3}
\end{array}
\right]\nonumber\\
{^{G}_{I_{i,k}}{\mathbf {\widetilde R}}}&= {^{G}_{I_{i,k}}{\mathbf {\hat R}}}({^{G}_{I_{i,k}}{\mathbf R}})^{\top} \nonumber\\
\bm\Gamma_{i,1}&={^G\hat{\mathbf v}}_{I_{i,k}} -{^{G}_{I_{i,k}}{\mathbf {\widetilde R}}} {^G{\mathbf v}_{I_{i,k}}} \nonumber\\
\bm\Gamma_{i,2}&={^G\hat{\mathbf p}}_{I_{i,k}} -{^{G}_{I_{i,k}}{\mathbf {\widetilde R}}} {^G{\mathbf p}_{I_{i,k}}} \nonumber\\
\bm\Gamma_{i,3}&={^G\hat{\mathbf p}}_{u} -{^{G}_{I_{i,k}}{\mathbf {\widetilde R}}} {^G{\mathbf p}_{u}}
\end{align}
$\widetilde{\mathbf B}_{i,k} \in \mathbb R^6$ is the error of the IMU bias given as
\begin{align}\label{eq_err_b}
\widetilde{\mathbf B}_{i,k}&=\begin{bmatrix}
(\hat{\mathbf b}_{\omega_{i,k}}-\mathbf b_{\omega_{i,k}})^\top & (\hat{\mathbf b}_{a_{i,k}}-\mathbf b_{a_{i,k}})^\top
\end{bmatrix}
\end{align}
and $\hat{\mathbf C}_{i,l} {\mathbf C}_{i,l}^{-1}$, for $l\in\{k-1,\cdots, k-m\}$,
is defined in $SE(3)$ as 

\begin{align}
\hat{\mathbf C}_{\!i,l} {\mathbf C}_{\!i,l}^{-1}&=
\left[
\scalebox{0.9}{$
\begin{array}{c@{\hspace{3pt}}c} 
{^{G}_{\!I_{i,l}}{\mathbf {\hat R}}}({^{G}_{\!I_{i,l}}{\mathbf R}})^{\top}&{^G\!\hat{\mathbf p}}_{\!I_{i,l}} -{^{G}_{\!I_{i,l}}{\mathbf {\hat R}}}({^{G}_{\!I_{i,l}}{\mathbf R}})^{\top} {^G{\!\mathbf p}_{\!I_{i,l}}} \\
\mathbf 0_{1\times3} & 1
\end{array}
$}
\right]
\end{align}

By applying the log-linear property of the invariant error \cite{7523335}, errors $\hat{\mathbf T}_{i,k} {\mathbf T}_{i,k}^{-1}$ and errors $\hat{\mathbf C}_{i,l} {\mathbf C}_{i,l}^{-1}$ can be approximated using a first-order approximation as follows:
\begin{align}\label{eq_logln}
\hat{\mathbf T}_{i,k} {\mathbf T}_{i,k}^{-1}&=\exp_\mathcal G(\bm \xi_{I_{i,k}})\approx \mathbf I_{6}+\bm {\xi}_{I_{i,k}}^\wedge\in\mathbb R^{6\times 6}\nonumber\\
\hat{\mathbf C}_{i,l} {\mathbf C}_{i,l}^{-1}&=\exp_\mathcal G(\bm \xi_{c_{i,l}})\approx \mathbf I_{6}+\bm {\xi}_{c_{i,l}}^\wedge\in\mathbb R^{6\times 6}
\end{align}
where $(\cdot)^\wedge: \mathbb R^{\text{dim} \mathfrak g}\to \mathfrak g$ be the linear map that transforms the error vector $\bm \xi_{I_{i,k}}$ and $\bm \xi_{c_{i,l}}$ defined in the Lie algebra to its corresponding matrix representation \cite{7523335} as
\begin{align}\label{eq_err_lie1}
\bm \xi_{I_{i,k}}&\triangleq\begin{bmatrix}
(\bm \xi_{\theta_{i,k}})^\top & (\bm \xi_{v_{i,k}})^\top & (\bm \xi_{p_{i,k}})^\top & (\bm\xi_{u_{k}})^\top
\end{bmatrix}^\top \in \mathbb R^{12}\nonumber\\
\bm \xi_{\theta_{i,k}}&=\widetilde{\bm \theta}_{i,k}\nonumber=\log({^{G}_{I_{i,k}}{\mathbf {\widetilde R}}})\in\mathbb R^3\\
\bm \xi_{v_{i,k}}&={^G\hat{\mathbf v}}_{I_{i,k}}-(\mathbf I_3+\lfloor \widetilde{\bm \theta}_{i,k}\times\rfloor)^G \mathbf v_{I_{i,k}}\in\mathbb R^3\nonumber\\
\bm \xi_{p_{i,k}}&={^G\hat{\mathbf p}}_{I_{i,k}}-(\mathbf I_3+\lfloor \widetilde{\bm \theta}_{i,k}\times\rfloor)^G \mathbf p_{I_{i,k}}\in\mathbb R^3 \nonumber\\
\bm \xi_{u_{i,k}}&={^G\hat{\mathbf p}}_{u}-(\mathbf I_3+\lfloor \widetilde{\bm \theta}_{i,k}\times\rfloor)^G \mathbf p_{u}\in\mathbb R^3,
\end{align}
and 
\begin{align}\label{eq_err_lie2}
\bm \xi_{c_{i,l}}&\triangleq\begin{bmatrix}
(\bm \xi_{\theta_{i,l}})^\top & (\bm \xi_{p_{i,l}})^\top 
\end{bmatrix}^\top \in \mathbb R^{6}\nonumber\\
\bm \xi_{\theta_{i,l}}&=\widetilde{\bm \theta}_{i,k}\nonumber=\log({^{G}_{I_{i,l}}{\mathbf {\widetilde R}}})\in\mathbb R^3\\
\bm \xi_{p_{i,l}}&={^G\hat{\mathbf p}}_{I_{i,l}}-(\mathbf I_3+\lfloor \widetilde{\bm \theta}_{i,l}\times\rfloor)^G \mathbf p_{I_{i,l}}\in\mathbb R^3 
\end{align}
Given the error definitions in \eqref{eq_err_b}, \eqref{eq_err_lie1}, and \eqref{eq_err_lie2}, we define the overall error state vector at timestep $k$ as
\begin{align}\label{eq_err}
\widetilde{\mathbf x}_{i,k}&\triangleq\begin{bmatrix}
\widetilde{\mathbf x}_{I_{i,k}}^\top &\widetilde{\mathbf x}_{C_{i}}^\top 
\end{bmatrix}^\top \in \mathbb R^{18+6m}\nonumber\\
\widetilde{\mathbf x}_{I_{i,k}}&=
\begin{bmatrix}
\bm \xi_{I_{i,k}}^\top & \widetilde{\mathbf B}_{i,k}^\top 
\end{bmatrix}^{\top} \nonumber\\
\widetilde{\mathbf x}_{C_{i}}& =\begin{bmatrix}
\bm\xi_{c_{i,k-1}}^\top & \cdots & \bm\xi_{c_{i,k-m}}^{\top}
\end{bmatrix}^{\top}
\end{align}
\subsection{System Model}
\noindent\textbf{System motion model:}
In the proposed DC-VIRO, we adopt the general 3-D rigid-body kinematics as the system motion model, given by:
\begin{align}\label{eq_imu_kn}
{^{G}_{I_{i,k}}\dot{\mathbf R}}&={^{G}_{I_{i,k}}{\mathbf R}}\lfloor {^{I_{i,k}}\bm\omega} \times\rfloor,\,
{^G{\dot{\mathbf v}}_{I_{i,k}}}={^{G}_{I_{i,k}}{\mathbf R}} ({^{I_{i,k}}\mathbf a})+{^G\mathbf g}\nonumber\\
{^G{\dot{\mathbf p}}_{\!I_{i,k}}}&{=}{^G{\mathbf v}_{\!I_{i,k}}},\,
{^G{\dot{\mathbf p}}_{u}}=\mathbf 0,\,
\dot{\mathbf b}_{\omega_{i,k}}=\mathbf w_{\!\omega_{i}},\, \dot{\mathbf b}_{\!a_{i,k}}=\mathbf w_{\!a_{i}}
\end{align}
where $^G \mathbf g=\begin{bmatrix}0&0&-9.8
\end{bmatrix}^\top$ denotes the gravity vector.
In particular, the IMU sensor can measure the robot's acceleration ${^{I_{i,k}}\mathbf a}$ and angular velocity ${^{I_{i,k}}\bm\omega}$ with respect to the IMU's own frame $I_{i,k}$ at each timestep $k$ as
\begin{align}
{{^{I_{i,k}}\mathbf a}_m}&={^{I_{i,k}}\mathbf a}-{^{G}_{I_{i,k}}{\mathbf R}}^\top {^G \mathbf g}+ {\mathbf n}_{a_{i,k}}+\mathbf b_{a_{i,k}}\nonumber\\
{^{I_{i,k}}\bm\omega}_m&={^{I_{i,k}}\bm\omega}+\mathbf n_{\omega_{i,k}}+\mathbf b_{\omega_{i,k}}
\end{align}
where ${{^{I_{i,k}}\mathbf a}_m}$ and ${^{I_{i,k}}\bm\omega}_m$ are the noisy measurements of the robot's acceleration and angular velocity, respectively.
$\mathbf n_{a_{i,k}}$ and $\mathbf n_{\omega_{i,k}}$ are the IMU's measurement noises assumed to be zero-mean Gaussian. $\mathbf b_{\omega_{i,k}}$ and $\mathbf b_{a_{i,k}}$ are the IMU's biases which are modeled as random walk process, where their time derivatives follow a Gaussian distribution.

\noindent\textbf{Camera measurement model:}
When a static landmark of the environment, denoted as ${^{G}\mathbf p_f}$, is tracked by the camera carried by robot $i$ at timestep $k$, the corresponding feature measurement can be obtained through the following model
\begin{align}\label{eq_cm}
\mathbf z_{c_{i,k}}&=\prod({^{C_{i,k}}\mathbf p_f})+n_{c_{i,k}}\nonumber\\
{^{C_{i,k}}\mathbf p_f}&={^{C_{i,k}}_{I_{i,k}} \mathbf R}{^{I_{i,k}}_G \mathbf R}({^{G}\mathbf p_f}-{^{G}\mathbf p_{I_{i,k}}})+{^{C_{i,k}}\mathbf p_{I_{i,k}}}
\end{align}
where $\mathbf n_{c_{i,k}}\sim \mathcal N(0, \mathbf Q_c)$ is the measurement noise which is assumed to be white Gaussian with covariance $\mathbf Q_c$. The ${^{C_{i,k}}\mathbf p_f}$ is the landmark position in the camera frame, and the projection function $\prod(\cdot)$ is defined as $\prod(\begin{bmatrix}x&y&z\end{bmatrix}^\top)=\begin{bmatrix}x/z&y/z
\end{bmatrix}^\top$

\noindent\textbf{Range measurement model:}
The UWB tag on the robot can provide the ranging measurement 
between the robot $i$ and the UWB anchor at time $t_k$, denoted as $d_{i,k}$. Given the robot's pose $({^{I_{i,k}}_G{\mathbf R}}, ^G\mathbf p_{I_{i,k}})$ and the anchor position ${^G\mathbf p_{u}}$, the ranging measurement is described by the following model:
\begin{align}\label{eq_uwb}
\mathbf z_{\!u_{i,k}}&{=}\|^G\mathbf p_{\!I_{i,k}}{+}{^{\!I_{i,k}}_G{\!\mathbf R}}^{\!\top}{^{\!I_{i,k}}\mathbf p_{T}}{-}{^{\!G}\mathbf p_{\!u}}\|{+}\mathbf n_{\!u_{i,k}}{+}\mathbf b_{\!u_{i,k}}
\end{align}
where $\mathbf n_{u_{i,k}}\sim \mathcal N(0, \mathbf Q_u)$ is the measurement noise; ${^{I_{i,k}}\mathbf p_T}$, $\mathbf b_{u_{i,k}}$ represent UWB tag's position in the IMU frame and the measurement bias, respectively. Note that the terms, ${^{I_{i,k}}\mathbf p_T}$ and $\mathbf b_{u_{i,k}}$ can be easily calibrated offline \cite{NTUviral}.

\section{Algorithm Design}
In this section, we present the proposed DC-VIRO algorithm for 3-D multi-robot systems, which enables simultaneous self-localization and calibration of multiple unknown anchors by leveraging inter-robot information as additional constraints. The main algorithm is summarized in Algorithm \ref{alg1}.

\subsection{IMU Propagation}
Consider a posterior state estimate $\hat{\mathbf X}_{i,k}$.\footnote{Throughout this paper, $(\bar{\mathbf X}_{i,k}, \bar{\mathbf P}_{i,k})$, $(\hat{\mathbf X}_{i,k}, \hat{\mathbf P}_{i,k})$ denotes the prior and posterior estimate of the robot $i$'s state $\mathbf X_{i,k}$ at timestep $t_{k}$, respectively.}
To propagate the state covariance, we first linearize the continuous-time kinematics \eqref{eq_imu_kn} at $\hat {\mathbf X}_{i,k}$ to compute a linearized error-state model as
\begin{align}\label{eq_lin}
\frac{d}{dt}{\widetilde{\mathbf x}}_{i,k}&=
\begin{bmatrix}
\mathbf F_{i,k} & \mathbf 0_{18\times 6m}\\
\mathbf 0_{6m\times18} & \mathbf 0_{6m}
\end{bmatrix}
{\widetilde{\mathbf x}}_{i,k}+
\begin{bmatrix}
\mathbf G_{i,k} \\ \mathbf 0_{6m\times18}
\end{bmatrix}\mathbf n_{i,k},
\end{align}
where $\mathbf n_{i,k}=\begin{bmatrix}
\mathbf n_{\omega_{i,k}}^\top & \mathbf n_{a_{i,k}}^\top & \mathbf 0_{1\times 6}^\top & \mathbf w_{\omega_{i}}^\top & \mathbf w_{a_{i}}^\top
\end{bmatrix}^\top$. The explicit forms of the Jacobians $\mathbf F_{i,k}$ and $\mathbf G_{i,k}$ are provided in Appendix.

Given the linearized system in \eqref{eq_lin}, we can propagate the covariance as 
\begin{align}
\hat{\mathbf P}_{i,k+1} &= \bar{\bm\Phi}_{i,k+1|k}\hat{\mathbf P}_{i,k}\bar{\bm\Phi}_{i,i,k+1|k}^\top + \mathbf Q_{i,k}
\end{align}
where $\mathbf Q_{i,k}$ denotes the noise covariance for the $i$'th robot. $\bar{\bm\Phi}_{i,k+1|k}$ is the discrete-time state transition matrix from time $t_{k}$ to $t_{k+1}$ which can be computed as
\begin{align}
    \bar{\bm\Phi}_{i,k+1|k}&=\exp\left( \begin{bmatrix}
\mathbf F_{i,k} & \mathbf 0_{18\times 6m}\\
\mathbf 0_{6m\times18} & \mathbf 0_{6m}
\end{bmatrix} \delta t\right) \nonumber\\
    \delta t &= t_{k+1} - t_{k}
\end{align}

\subsection{Range Update}
We first linearize the range measurement model \eqref{eq_uwb} by using the first-order Taylor expansion. Notably, unlike the standard Taylor expansion represented with the standard vector error, we have to linearize \eqref{eq_uwb} with the right-invariant error $\widetilde{\mathbf x}_{I_{i,k}}$ in \eqref{eq_err}. Recall that $\exp_{\mathcal G}(\bm \xi_{I_{i,k}})\approx \mathbf I_6 +\bm \xi_{I_{i,k}}^\wedge$ 
in \eqref{eq_logln}, we have the following linearized model at the linearization point $\hat{\mathbf X}_{i,k}$ as
\begin{align}\label{eq_ln_uwb}
\widetilde{\mathbf z}_{u_{i,k}}&= \mathbf H_{u_{i,k}} \widetilde{\mathbf x}_{I_{i,k}} + \mathbf n_{u_{i,k}}
\end{align}
where the measurement Jacobian $\mathbf H_{u_{i,k}}$ is given by
\begin{align}
\mathbf H_{u_{i,k}}&= \mathbf H_{pu}\begin{bmatrix}
\Lambda_k & \mathbf 0_3 & \mathbf I_3 & -\mathbf I_3 & \mathbf 0_{3\times 6}
\end{bmatrix}\nonumber\\
\mathbf H_{pu}&=\frac{ \left({^G \hat{\mathbf p}_{I_{i,k}}}-{^G \hat{\mathbf p}_u}+{^G_{I_{i,k}} \hat{ \mathbf R}}\, {^{I_{i,k}}\mathbf p_T} \right)^\top}{\|  {^G \hat{\mathbf p}_{I_{i,k}}}-{^G \hat{\mathbf p}_u}+{^G_{I_{i,k}} \hat{ \mathbf R}}\, {^{I_{i,k}}\mathbf p_T}\|} \nonumber\\
\Lambda_k&= \lfloor{ {^G \hat{\mathbf p}_u}-{^G \hat{\mathbf p}_{I_{i,k}}}-{^G_{I_{i,k}} \hat{ \mathbf R}}\, {^{I_{i,k}}\mathbf p_T}\times\rfloor}.
\end{align}

As previously mentioned, UWB anchors can be classified into two categories: individual anchors, which are observed by only a single robot, and common anchors, which can be simultaneously observed by multiple robots. For measurements associated with individual anchors, one can directly incorporate them into the update step following the standard Invariant EKF framework~\cite{YY2022}. Therefore, this work focuses on the update with common anchor measurements.

Specifically, if both robot $i$ and its neighbors receive distance measurements from the same anchor, this indicates that the anchor is a common anchor at this time step. Let $\mathcal N_{i,k}=\{j_1,\ldots,j_L\}$ denote the neighbors of robot $i$ that observe the common anchor except robot $i$, given as
$\mathcal N_{i,k}=\{j_1,\ldots,j_L\}$. To update using this common anchor information, each robot $i$ will collect information from their neighbor to compute a stacked linearized system as
\begin{align}\label{eq_stack}
\bar{\mathbf z}_{u_i}=
\begin{bmatrix}
\widetilde{\mathbf z}_{u_{i,k}}\\\widetilde{\mathbf z}_{u_{j_{1,k}}}\\\vdots\\\widetilde{\mathbf z}_{u_{j_{L,k}}}
\end{bmatrix} 
=  
&\widetilde{\mathbf H}_{i,k}
\begin{bmatrix} 
\widetilde{\mathbf x}_{I_{i,k}}\\\widetilde{\mathbf x}_{I_{j_{1,k}}}\\\vdots\\\widetilde{\mathbf x}_{I_{j_{L,k}}}
\end{bmatrix} +\begin{bmatrix} 
\mathbf n_{u_{i,k}} \\ \mathbf n_{u_{j_{1,k}}} \\ \vdots \\ \mathbf n_{u_{j_{L,k}}}
\end{bmatrix}
\end{align}
where $\widetilde{\mathbf H}_{i,k}=\text{diag}(\mathbf H_{u_{i,k}}, \mathbf H_{u_{j_{1,k}}}, \dots, \mathbf H_{u_{j_{L,k}}})$ denotes the measurement jacobian of the stacked system. 
For simplicity, we reformulate model \eqref{eq_stack}, where each column of $\widetilde{\mathbf H}_{i,k}$ is rewritten as a block matrix:
\begin{align}\label{eq_stack_simp}
\scalebox{0.9}{$
\begin{bmatrix}
\widetilde{\mathbf z}_{u_{i,k}}\\\widetilde{\mathbf z}_{u_{j_{1,k}}}\\\vdots\\\widetilde{\mathbf z}_{u_{j_{L,k}}}
\end{bmatrix}$}&\triangleq 
\scalebox{0.9}{$
\begin{bmatrix}
\bar{\mathbf H}_{u_{i,k}}& \dots& \bar{\mathbf H}_{u_{j_{L,k}}}    
\end{bmatrix}
$}
\scalebox{0.9}{$
\begin{bmatrix} 
\widetilde{\mathbf x}_{I_{i,k}}\\\widetilde{\mathbf x}_{I_{j_{1,k}}}\\\vdots\\\widetilde{\mathbf x}_{I_{j_{L,k}}}
\end{bmatrix}$} + \scalebox{0.9}{$\begin{bmatrix} 
\mathbf n_{u_{i,k}} \\ \mathbf n_{u_{j_{1,k}}} \\ \vdots \\ \mathbf n_{u_{j_{L,k}}}
\end{bmatrix}$}
\end{align}

This stacked measurement model depends on the states of all robots in $\mathcal N_{i,k}$, thereby incorporating information from both the ego-robot and its neighbors, and effectively improving estimation accuracy by leveraging shared inter-robot measurements. However, it is important to note that this stacked measurement model cannot be directly used for the update, since states $\widetilde{\mathbf x}_{I_{i,k}}$ and $\widetilde{\mathbf x}_{I_{j_{l,k}}}$ become correlated once robots utilize shared information \cite{CVIO}. Such cross-correlation is generally unknown in a fully distributed setting, and if ignored, it may lead to inconsistency and degraded estimation performance \cite{CVIO}. To properly handle the unknown correlation term and ensure the consistency of the estimator, we leverage the covariance intersection algorithm \cite{CVIO} to approximate the covariance
\begin{align}\label{eq_ci}
\begin{bmatrix}
\frac{1}{\omega_i} \bar{\mathbf P}_i & \\
& \ddots \\
& &\frac{1}{\omega_{j_L}} \bar{\mathbf P}_{j_L}
\end{bmatrix}
>
\begin{bmatrix}
\bar{\mathbf P}_{i} & \cdots & \bar{\mathbf P}_{ij_L} \\
\vdots & \ddots & \vdots \\
\bar{\mathbf P}_{ij_L}^\top & \cdots & \bar{\mathbf P}_{j_L}
\end{bmatrix}
\end{align}
where the right-hand side denotes the true covariance matrix that includes the cross-covariance terms, while the left-hand side represents the approximated covariance matrix expressed as a function of the individual covariances of each robot. The coefficients $\omega_i$ and $\omega_{j_l}$ are fusion weights that satisfy
\begin{align}
\omega_i>0, \omega_{j_l}>0, \omega_i+\sum_{j_l\in\mathcal N_{i,k}}\omega_{j_l}=1
\end{align}
Using the linearized measurement residual in \eqref{eq_stack_simp}, the EKF update for the state of each robot can be performed by substituting the approximated covariance in Equation \eqref{eq_ci}. In particular, the covariance can be updated as
\begin{align}
\hat{\mathbf P}_{i,k} &= \frac{1}{\omega_i} \bar{\mathbf P}_{i,k} - \frac{1}{\omega_i^2} \bar{\mathbf P}_{i,k} \bar{\mathbf H}^{\top}_{u_{i,k}} \mathbf{S}_k^{-1} \bar{\mathbf H}_{u_{i,k}} \bar{\mathbf P}_{i,k}\nonumber\\
\mathbf{S}_{i,k} &= \sum_{j \in \mathcal{N}_{i,k}} \frac{1}{\omega_j} \bar{\mathbf H}_{u_{j,k}} \bar{\mathbf P}_{j,k} \bar{\mathbf H}_{u_{j,k}}^{\top} + \mathbf{Q}_{u}
\end{align}
and the state is updated as
\begin{align}\label{eq_up}
\hat {\mathbf X}_{i,k}&= \exp(\bm \varepsilon_{i,k})\bar {\mathbf X}_{i,k}
\nonumber\\
\bm \varepsilon_{i,k} &= \frac{1}{\omega_i} \bar{\mathbf P}_{i,k} \bar{\mathbf H}^{\top}_{u_{i,k}} \mathbf{S}_k^{-1} \bar{\mathbf z}_{u_i}
\end{align}

\subsection{Visual Update}
Similar to the UWB measurement update, the visual update is also divided into two categories: \emph{common observation features} and \emph{individual observation features}. The term \emph{common observation features} refers to landmarks or features that are simultaneously observed by multiple robots. These shared observations can be exploited to establish inter-robot constraints and improve the consistency of the joint state estimation. In contrast, \emph{individual observation features} are those detected by only a single robot, and thus contribute solely to the local state update of that robot without introducing cross-robot constraints. Specifically, we first linearize the 
\begin{align}\label{eq_cm_up}
\widetilde{\mathbf z}_{c_{i,k}} = \mathbf H_{x_{i,k}} \widetilde{\mathbf x}_{I_{i,k}} + \mathbf H_{f_{i,k}}\widetilde{\mathbf p}_f + \mathbf n_{c_{i,k}}
\end{align}
where the measurement jacobians $\mathbf H_{x_{i,k}}$ and $\mathbf H_{f_{i,k}}$ can be computed as
\begin{align}
\mathbf H_{x_{i,k}}&= \mathbf H_{pc}{^{C_{i,k}}_{I}\hat{\mathbf R}}{^{I_{i,k}}_G\hat{\mathbf R}}\begin{bmatrix} \mathbf 0_3 & \mathbf 0_3 & -\mathbf I_3 & \mathbf 0_{3\times 9}
\end{bmatrix}\nonumber\\
\mathbf H_{f_{i,k}}&=\mathbf H_{pc}{^{C_{i,k}}_{I_{i,k}}\hat{\mathbf R}}{^{I_{i,k}}_G\hat{\mathbf R}}, \,
\mathbf H_{pc}=\left[\begin{array}{@{\!}c@{\;}c@{\;}c@{\!}}
1/\hat{z}& 0 & -\hat x/\hat{z}^2\\
0 & 1/\hat{z} & -\hat y/\hat{z}^2
\end{array}.
\right]
\end{align}
For individual observation features, the linearized measurement residual in \eqref{eq_cm_up} can be directly employed for the update. In contrast, when a common observation feature $^G \mathbf p_f$ is detected by multiple robots, each robot $i$ aggregates the information received from its neighbors and constructs a stacked linearized system as
\begin{align}\label{eq_stack_1}
\bar{\mathbf z}_{c_i}=
\scalebox{0.9}{$
\begin{bmatrix}
\widetilde{\mathbf z}_{c_{i,k}}\\\widetilde{\mathbf z}_{c_{j_{1,k}}}\\\vdots\\\widetilde{\mathbf z}_{c_{j_{L,k}}}
\end{bmatrix}
$}
=  
&\bar{\mathbf H}_{i,k}
\scalebox{0.9}{$
\begin{bmatrix} 
\widetilde{\mathbf x}_{I_{i,k}}\\\widetilde{\mathbf x}_{I_{j_{1,k}}}\\\vdots\\\widetilde{\mathbf x}_{I_{j_{L,k}}}
\end{bmatrix}$} + \bar{\mathbf H}_{f_i}{^G\bar{\mathbf p}}_f+
\scalebox{0.9}{$
\begin{bmatrix} 
\mathbf n_{c_{i,k}} \\ \mathbf n_{c_{j_{1,k}}} \\ \vdots \\ \mathbf n_{c_{j_{L,k}}}
\end{bmatrix}
$}
\end{align}
where $\bar{\mathbf H}_{i,k}=\text{diag}(\mathbf H_{x_{i,k}}, \mathbf H_{x_{j_{1,k}}}, \dots, \mathbf H_{x_{j_{L,k}}})$ denotes the measurement jacobian with respect to all involved robots' states, $\bar{\mathbf H}_{f_i}=\begin{bmatrix}
\mathbf H_{f_{i,k}}^\top& \mathbf H_{f_{j_{1,k}}}^\top& \dots& \mathbf H_{f_{j_{L,k}}}^\top
\end{bmatrix}^\top$ is the jacobian with respect to the feature point. Since it follows the same form as \eqref{eq_stack_simp}, the procedures in \eqref{eq_stack_simp}–\eqref{eq_up} can subsequently be employed to perform the update with common observation features.

\subsection{UWB state augmentation and initialization}
Initially, the system state estimate $\hat{\mathbf{T}}_k$ only includes the IMU state and not the anchor position $^G\hat{\mathbf{p}}_u$. Thus, a key task is to initialize the UWB anchor state together with its covariance and augment them into the system state, enabling joint estimation within the DC-VIRO framework. We assume that each robot can directly obtain IMU states and ranging measurements over a time window of length $n$, and can share its ranging information with neighboring robots within the UWB measurement range. Specifically, to initialize the anchor state, we construct the following optimization problem based on ranging measurements for each robot $i$:
\begin{align}
\min_{^G\mathbf{p}_{u_i}}\sum_{k=1}^{n}(\mathbf{z}_{u_{i,k}} - \mathbf{b}_{u_{i,k}} - \left\|{^{G}\mathbf{p}_{T_{i,k}}} - {^{G}\mathbf{p}_{u_{i}}} \right\|)^2
\end{align}
where $^G\mathbf{p}_{T_{i,k}} = ^G\mathbf{p}_{I_{i,k}} + ^{I_{i,k}}_{G} \mathbf{R}^{\top} {^{I_{i,k}}\mathbf{p}_{T}}$ denotes the position of the UWB tag carried by the $i$-th robot relative to the IMU frame $I_{i,k}$. 

To initialize the covariance of UWB anchor states and their correlation with existing states, we first define the state $\mathbf{X}_{i_{pos}}$ containing the pose of the $i$-th robot from time step $k=0$ to $k=n$.
\begin{align}
\mathbf{X}_{i_{pos}} = \begin{bmatrix}{^G_{I_{i,0}}\mathbf{R}^{\top}} & {^G\mathbf{p}_{I_{i,0}}^\top} & \cdots & {^{G}_{I_{i,n}}\mathbf{R}^{\top}} &{^{G}\mathbf{p}_{I_{i,n}}^\top}\end{bmatrix}^{\top}
\end{align}

Then stack all available UWB measurements $\mathbf{z}_{u,i}=\begin{bmatrix}\mathbf{z}_{u_{0},i},\cdots,\mathbf{z}_{u_{n},i}\end{bmatrix}$ to construct the following stacked measurement model:
\begin{align}
\mathbf{z}_{u,i} = \mathbf{h}\left(\mathbf{X}_{i_{pos}},^{G}\mathbf{p}_{u}\right) + \mathbf{n}_{u,i}
\end{align}
Each $\mathbf{z}_{u_{i}}$ satisfies model \eqref{eq_uwb}. Linearizing the measurement model above, we obtain:
\begin{align}
\widetilde{\mathbf{z}}_{u,i} = \begin{bmatrix}\mathbf{H}_{x,i} & \mathbf{H}_{u,i}\end{bmatrix} \begin{bmatrix}\widetilde{\mathbf{x}}_{i_{pos}} \\ \mathbf{\xi}_{u_k} \end{bmatrix} + \mathbf{n}_{u,i}
\end{align}
$\mathbf{n}_{u,i}$ represents the stacked measurement noise vector, $\widetilde{\mathbf{x}}_{i_{pos}}$ represents the error state of $\mathbf{X}_{i_{pos}}$ given by:
\begin{align}
\widetilde{\mathbf{x}}_{i_{pos}} = \begin{bmatrix} \bm{\xi}_{\theta_{i,0}}^{\top} & \bm{\xi}_{p_{i,0}}^{\top} & \cdots & \bm{\xi}_{\theta_{i,n}}^{\top} & \bm{\xi}_{p_{i,n}}^{\top}\end{bmatrix}^{\top}
\end{align}
Through QR decomposition \cite{geneva2020openvins}, we decompose the linear system into two subsystems:
\begin{align}
\begin{bmatrix}\widetilde{\mathbf{z}}_{u_{1,i}} \\ \widetilde{\mathbf{z}}_{u_{2,i}}\end{bmatrix} = \begin{bmatrix}\mathbf{H}_{x_{1,i}} & \mathbf{H}_{u_{1,i}} \\ \mathbf{H}_{x_{2,i}} & 0\end{bmatrix} \begin{bmatrix}\widetilde{\mathbf{x}}_{i_{pos}} \\ \mathbf{\xi}_{u_k} \end{bmatrix} + \begin{bmatrix} \mathbf{n}_{u_{1,i}} \\ \mathbf{n}_{u_{2,i}}\end{bmatrix}
\end{align}
Then, the covariance of the UWB state and its correlations with the existing state can be computed and incorporated into the current covariance through augmentation, following \cite{geneva2020openvins}.

\begin{algorithm}[h]
\caption{DC-VIRO}
\label{alg1}
\begin{algorithmic}[1]
\State \textbf{Step 1: Initialization}
\State \quad Consider the nonlinear system motion model \eqref{eq_imu_kn}. Start with $\mathbf{\hat X}_{i,0} = \mathbb{E}(\mathbf X_{i,0})$, $\mathbf{\hat P}_{i,0} = \mathbf{P}_{i,0}$, for all robots $i=1,...,n$.
\State \textbf{Step 2: State propagation by each single robot}
\State \quad For each agent $i=1,...,n$, performs the propagation step to compute the prior estimate $(\mathbf{\bar X}_{i,k}, \mathbf{\bar P}_{i,k})$ at timestep $k$.
\State \textbf{Step 3: State augmentation}
Once robot $i$ detects a previously unseen UWB anchor, it initializes the corresponding anchor state and augments both the state and its covariance into the overall system state.
\State \textbf{Step 4: Measurement update}
\State \quad Each robot communicates with its neighbors to determine whether commonly observed features or shared anchors exist. If no common observations are available, each robot performs the update step using only its own measurements. When common features or anchors are detected, the shared information is fused to compute the posterior estimate $(\hat{\mathbf{X}}_{i,k}, \hat{\mathbf{P}}_{i,k})$.
\end{algorithmic}
\end{algorithm}

\section{Consistency Analysis}
\subsection{Observability Analysis}\label{ob_ana}
Observability is a crucial metric for analyzing system consistency. In this section, we conduct an observability analysis of the DC-VIRO system constructed above. To simplify the derivation, we assume that each robot's state contains only one feature point $^G\mathbf{p}_{f}$ and one UWB anchor point $^G\mathbf{p}_{u}$.
\begin{align}
\mathbf{T}_{\!i,k} = \left[ \begin{array}{c|c}
^G_{\!I_{i,k}}\mathbf{R} & 
^G\mathbf{v}_{\!I_{i,k}} \, ^G\mathbf{p}_{\!I_{i,k}}\, ^G\mathbf{p}_{\!u} \, ^G\mathbf{p}_{\!f} \\
\hline
\mathbf{0}_{3\times4} &\mathbf{I}_4 \\
\end{array} \right] \in SE_4(3)
\end{align}
According to the proof in reference \cite{huang2009first}, the local observability matrix of the time-varying error states for the entire system is defined as:
\begin{align}
\mathcal{O} = \begin{bmatrix} \mathcal{O}_0 \\ \mathcal{O}_1 \\ \vdots \\ \mathcal{O}_k \end{bmatrix} = \begin{bmatrix} \mathbf{H}_0 \\ \mathbf{H}_1 {\bar{\boldsymbol{\Phi}}}_{1|0} \\ \vdots \\ \mathbf{H}_k {\bar{\boldsymbol{\Phi}}}_{k|0} \end{bmatrix},
\end{align}
Here, $\bar{\bm\Phi}_{k|0} = \text{diag}({\bm\Phi}_{1,k|0},\cdots,{\bm\Phi}_{n,k|0})$ is the joint state transition matrix for all robots. Since we use visual measurements and distance measurements to update the state, the joint measurement Jacobian matrix can be written as 
\begin{align}
\mathbf{H}_{k} = 
\begin{bmatrix}
\mathbf{H}_{c_{k}}^{\top}&\mathbf{H}_{u_{k}}^{\top}      
\end{bmatrix}^\top
\end{align}
where $\mathbf{H}_{c_{k}} =\text{diag}(\mathbf{H}_{c_{1,k}},\cdots,\mathbf{H}_{c_{n,k}})$ represents the Jacobian matrix for visual measurements, and $\mathbf{H}_{u_{k}} =\text{diag}(\mathbf{H}_{u_{1,k}},\cdots,\mathbf{H}_{u_{n,k}})$ denotes the Jacobian matrix for UWB ranging measurements. The $k$-th sub-block of the observability matrix can then be expressed as
\begin{align}
\mathcal{O}_k \triangleq
\begin{bmatrix}
\mathcal{O}_{c_{k}} \\
\mathcal{O}_{u_{k}}
\end{bmatrix}
= \begin{bmatrix}
\mathbf{H}_{c_{k}} {\bar{\boldsymbol{\Phi}}}_{k|0} \\
\mathbf{H}_{u_{k}} {\bar{\boldsymbol{\Phi}}}_{k|0}
\end{bmatrix}
\end{align}
where each sub-block can be represented as
\begin{align}
\mathbf{H}_{c_{k}} \bar{\boldsymbol{\Phi}}_{k|0} =
\begin{bmatrix} \mathbf{H}_{c_{1,k}} {\boldsymbol{\Phi}}_{1,k|0} \\ &\ddots \\ &&\mathbf{H}_{c_{n,k}}\boldsymbol{\Phi}_{n,k|0} \end{bmatrix}\nonumber\\
\mathbf{H}_{u_{k}} \bar{\boldsymbol{\Phi}}_{k|0} =
\begin{bmatrix} \mathbf{H}_{u_{1,k}} {\boldsymbol{\Phi}}_{1,k|0} \\ &\ddots \\ &&\mathbf{H}_{u_{n,k}}\boldsymbol{\Phi}_{n,k|0} \end{bmatrix}
\end{align}

\begin{lemma}
    If each robot in the team has general motion in the 3-D space, the right null space $\mathbf{N}$ of the observability matrix $\mathcal{O}_{k}$ for a multi-robot system can be composed of four unobservable directions corresponding to global position and yaw rotation:
    \begin{align}
    \mathbf{N} = \begin{bmatrix}
    \mathbf{N_{1}} \\
    \vdots \\
    \mathbf{N_{n}}
    \end{bmatrix},\quad
    \mathbf{N}_i = \begin{bmatrix}
    ^G\mathbf{g} & \mathbf{0}_3 \\
    \mathbf{0}_{3\times 1} & \mathbf{0}_3\\
    \mathbf{0}_{3\times 1} & \mathbf{I}_3 \\
    \mathbf{0}_{3\times 1} & \mathbf{I}_3 \\
    \mathbf{0}_{3\times 1} & \mathbf{I}_3 \\
    \mathbf{0}_{3\times 1} & \mathbf{0}_3 \\
    \mathbf{0}_{3\times 1} & \mathbf{0}_3
    \end{bmatrix}, \quad \text{for } i = \{1, \dots, n\}
    \end{align}
\end{lemma}

\begin{remark}
Owing to the invariant error property, the unobservable subspace of the original multi-robot system remains independent of the state estimate. Consequently, the proposed DC-VIRO algorithm inherits the same observability characteristics as the original system, thereby mitigating the inconsistency introduced by linearization in standard EKF and ensuring estimator consistency.
\end{remark}

\section{Simulations}
\subsection{Simulation Settings}
\begin{figure}[h]
    \centering
    \begin{subfigure}[b]{0.23\textwidth}
        \centering
        \includegraphics[width=\textwidth]{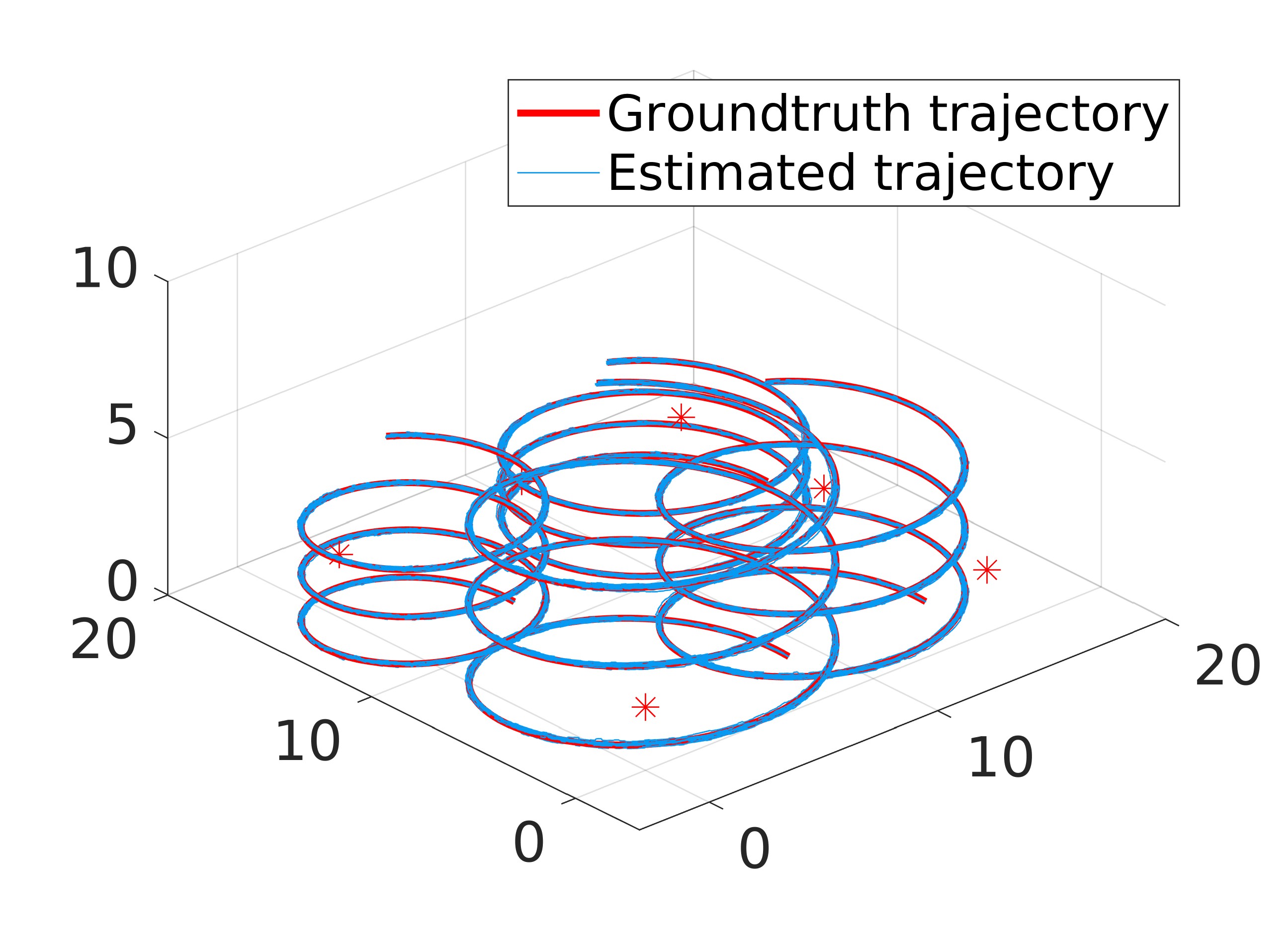}
        \caption{Trajectory a}
        \label{fig:subfig1}
    \end{subfigure}
    \hfill
    \begin{subfigure}[b]{0.23\textwidth}
        \centering
        \includegraphics[width=\textwidth]{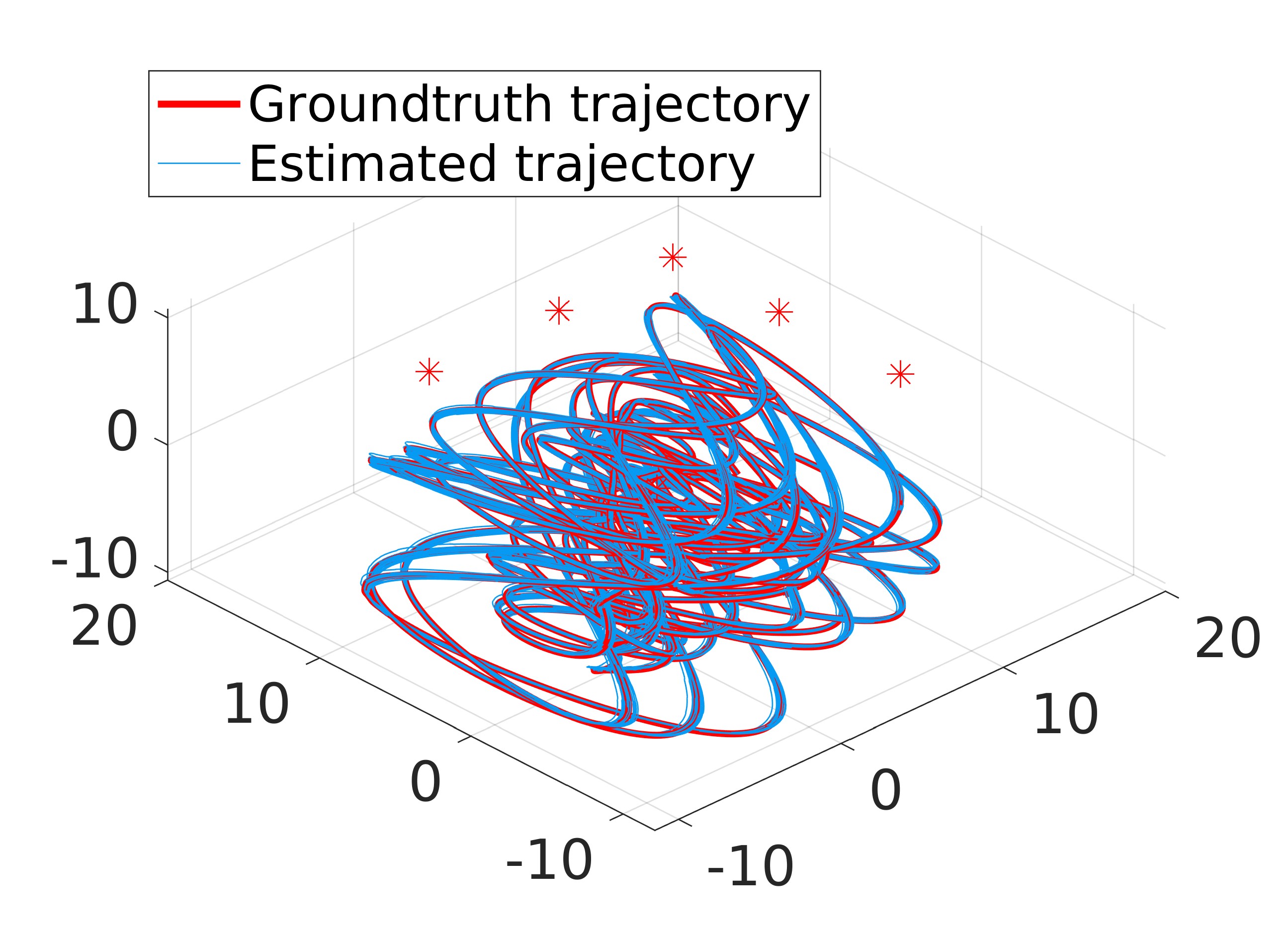}
        \caption{Trajectory b}
        \label{fig:subfig2}
    \end{subfigure}
    \caption{Robots' groundtruth path and the estimation performance over 100 Monte-Carlo simulation under different trajectories. }
    \label{fig_traj}
\end{figure}

We used MATLAB to simulate a three-dimensional environment with multiple robots and $3$ UWB anchors, where each robot follows a predefined trajectory, as illustrated in Fig.~\ref{fig_traj}. Each robot is equipped with an IMU for motion sensing, a monocular camera for feature detection, and a UWB tag for ranging with anchors. The IMU, monocular camera, and UWB measurements operate at frequencies of $100$ Hz, $10$ Hz, and $10$ Hz, respectively. The IMU noise standard deviations for each robot are specified as
\begin{align}
    \mathbf n_{a_i}&=\left[0.003,0.003,0.004\right]^{\top}(m/(s^{2}\sqrt{Hz}))\nonumber\\
\mathbf n_{\omega_i}&=\left[0.0003,0.0003,0.0005\right]^{\top}(rad/(s^{2}\sqrt{Hz}))
\end{align}

The camera has a limited observation range, and its measurement noise is modeled as $1$ pixel. The initial positions of the UWB anchors are unknown and must be calibrated, with a noise level of $0.1 m$ . The anchor positions are listed in Table~\ref{tab:anchor_init}. The initial covariance of each robot’s state (position, velocity, and orientation) is set to $10^{-3}\mathbf{I}_{9}$, and the initial state estimates are initialized to their true poses, as summarized in Table~\ref{tab:pos_init}.

\begin{table}[h]
    \centering
    \caption{Initial pose of each anchor}
    \begin{tabular}{cc}
        \toprule
        Anchor & Position (m) \\
        \midrule
        1 & $\begin{bmatrix} 0.00 & 0.00 & 0.00 \end{bmatrix}^\top$ \\
        2 & $\begin{bmatrix} 0.00 & 15.00 & 2.00 \end{bmatrix}^\top$ \\
        3 & $\begin{bmatrix} 5.00 & 15.00 & 2.00 \end{bmatrix}^\top$ \\
        \bottomrule
    \end{tabular}
    \label{tab:anchor_init}
\end{table}

Each robot achieves joint information acquisition by communicating with neighboring robots in pairs, with the probability of establishing communication with neighbors set at $70\%$.

\begin{table}[h]
    \centering
    \caption{Initial pose of each robot for trajectories a}
    \begin{tabular}{ccc}
        \toprule
        Robot & Position (m) & Orientation (rad) \\
        \midrule
        1 & $\begin{bmatrix} 14.00 & 4.00 & 0.00 \end{bmatrix}^\top$ & $\begin{bmatrix} 0 & 0 & 0 \end{bmatrix}^\top$ \\
        2 & $\begin{bmatrix} 14.00 & 11.00 & 0.00 \end{bmatrix}^\top$ & $\begin{bmatrix} 0 & \pi & 0 \end{bmatrix}^\top$ \\
        3 & $\begin{bmatrix} 6.00 & 4.00 & 0.00 \end{bmatrix}^\top$ & $\begin{bmatrix} 0 & 0 & 0 \end{bmatrix}^\top$ \\
        4 & $\begin{bmatrix} 6.00 & 11.00 & 0.00 \end{bmatrix}^\top$ & $\begin{bmatrix} 0 & \pi & 0 \end{bmatrix}^\top$ \\
        \bottomrule
    \end{tabular}
    \label{tab:pos_init}
\end{table}

\subsection{Simulation Results}
\begin{figure}[h]
    \centering
    \begin{subfigure}[b]{0.44\textwidth}
        \centering
        \includegraphics[width=\textwidth]{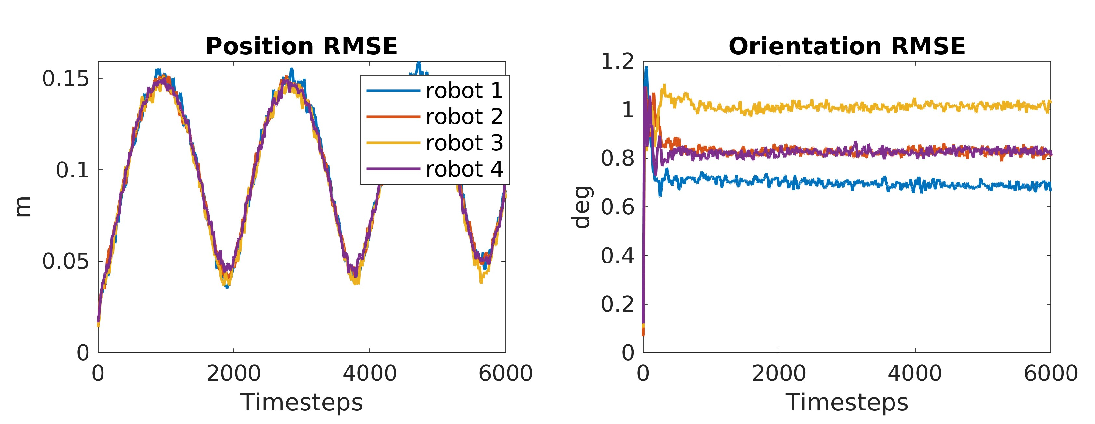}
        \caption{RMSE of Trajectory a}
        \label{fig:subfig1}
    \end{subfigure}
    \hfill
    \begin{subfigure}[b]{0.44\textwidth}
        \centering
        \includegraphics[width=\textwidth]{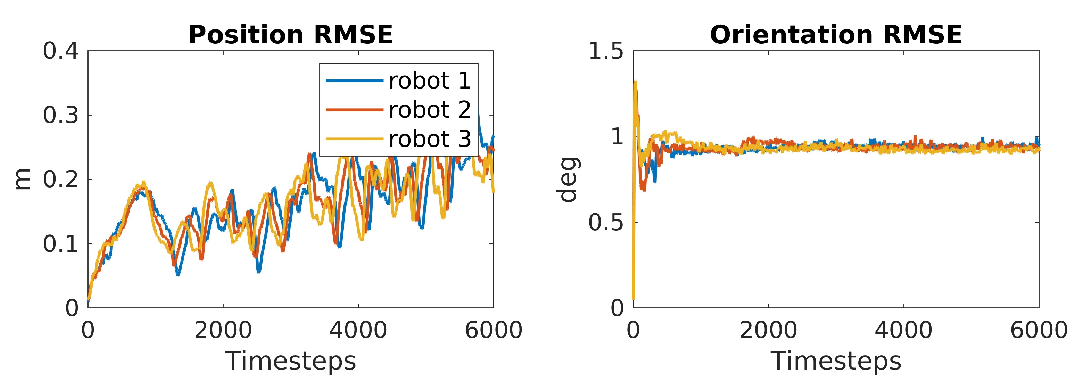}
        \caption{RMSE of Trajectory b}
        \label{fig:subfig2}
    \end{subfigure}
    \caption{Averaged position RMSE (PRMSE) and orientation RMSE (ORMSE) for each robot over 100 Monte-Carlo simulations.}
    \label{fig_rmse}
\end{figure}

To thoroughly evaluate the performance of the proposed algorithm, we conducted $100$ Monte Carlo simulation runs. The estimated trajectories of the robots are shown in Fig.~\ref{fig_traj}, demonstrating close agreement with the ground truth. Using the Root Mean Square Error (RMSE) as the evaluation metric, we assessed the accuracy of both position and heading estimation. The results, summarized in Fig.~\ref{fig_rmse}, present the average Position RMSE (PRMSE) and Orientation RMSE (ORMSE) for each robot, confirming that the proposed algorithm achieves highly accurate and consistent localization performance.

To further demonstrate its effectiveness, we compared the proposed algorithm with a single-robot localization approach that does not involve inter-robot information sharing. As reported in Table~\ref{tab:rmse}, incorporating inter-robot information fusion leads to a clear improvement in the overall localization accuracy of the robot team, confirming the advantage of the proposed method.
\begin{table}[h]
    \centering
    \caption{Comparison of average RMSE (m/deg) with and without inter-robot information}
    \begin{tabular}{ccc}
        \toprule
        & Trajectory a & Trajectory b \\
        \midrule
        w/ inter-robot info. & 0.147/1.295 & 0.256/1.027 \\
        w/o inter-robot info. & 0.205/1.788 & 0.498/1.575 \\
        \bottomrule
    \end{tabular}
    \label{tab:rmse}
\end{table}


\section{Conclusion}
This paper presents a consistent and tightly coupled distributed multi-robot visual–inertial–ranging odometry (DC-VIRO) system formulated on matrix Lie groups. By incorporating UWB ranging measurements into a collaborative VIO framework, the system introduces the concept of shared anchors, which, together with common visual features, provide additional geometric constraints through inter-robot communication. Anchor positions are explicitly included in the state to handle calibration uncertainty, and the use of a right-invariant error formulation ensures that DC-VIRO preserves the same four unobservable directions as standard VIO, guaranteeing observability consistency. Extensive Monte Carlo simulations demonstrate that the proposed method significantly improves multi-robot localization accuracy and robustness. In future work, we will focus on implementing DC-VIRO on real robotic platforms, addressing challenges such as sensor synchronization, communication delays, and real-time computation to validate its effectiveness in practical deployments.

\appendix
The Jacobians $\mathbf F_{i,k}$ and $\mathbf G_{i,k}$ are obtained by linearizing the continuous-time IMU kinematics in \eqref{eq_imu_kn} under the right-invariant error definition.
The Jacobian $\mathbf F_{i,k}$ can be computed as
\begin{align}
\mathbf F_{i,k}&=
\begin{bmatrix}
\mathbf F_A & \mathbf F_{B_{i,k}}\\
\mathbf 0_{6\times 12} & \mathbf 0_{6}
\end{bmatrix},\mkern8mu
\mathbf F_{A}=\begin{bmatrix}
\mathbf 0_3 & \mathbf 0_3 & \mathbf 0_3 & \mathbf 0_3\\
\lfloor^G\mathbf{g}\times\rfloor & \mathbf 0_3 & \mathbf 0_3 & \mathbf 0_3 \\
\mathbf 0_3 & \mathbf{I}_3 & \mathbf 0_3 & \mathbf 0_3 \\
\mathbf 0_3 & \mathbf 0_3 & \mathbf 0_3 & \mathbf 0_3 \\
\end{bmatrix}\nonumber\\
\mathbf F_{B_{i,k}}&=\begin{bmatrix}
-{^G_{I_{i,k}} \hat{\mathbf R}} & \mathbf 0_3\\
-\lfloor{^G\hat{\mathbf v}_{I_{i,k}} \times}\rfloor{^G_{I_{i,k}} \hat{\mathbf R}} & -{^G_{I_{i,k}} \hat{\mathbf R}} \\
-\lfloor{^G\hat{\mathbf p}_{I_{i,k}} \times}\rfloor{^G_{I_{i,k}} \hat{\mathbf R}} & \mathbf 0_3 \\
-\lfloor{^G\hat{\mathbf p}_{u} \times}\rfloor{^G_{I_{i,k}} \hat{\mathbf R}} & \mathbf 0_3
\end{bmatrix}
\end{align}
and the Jacobian $\mathbf G_{i,k}$ can be computed as
\begin{align}
\mathbf G_{i,k}&=
\begin{bmatrix}
\mathbf{Ad}_{\hat {\mathbf X}_{i,k}} & \mathbf 0_{12\times 6}\\
\mathbf 0_{6\times 12} & \mathbf 0_6
\end{bmatrix}\nonumber\\
\mathbf{Ad}_{\hat {\mathbf X}_{i,k}}&=
\scalebox{0.9}{$
\begin{bmatrix}
{^{G}_{I_{i,k}}\hat{\mathbf R}}& \mathbf 0_3 & \mathbf 0_3&\mathbf 0_3 \\
\lfloor{^G\hat{\mathbf v}_{I_{i,k}}\times \rfloor{^{G}_{I_{i,k}}}\hat{\mathbf R}}& {^{G}_{I_{i,k}}\hat{\mathbf R}} &\mathbf 0_3 & \mathbf 0_3 \\
\lfloor{^G\hat{\mathbf p}_{I_{i,k}}\times \rfloor{^{G}_{I_{i,k}}}\hat{\mathbf R}}& \mathbf 0_3& {^{G}_{I_{i,k}}\hat{\mathbf R}} & \mathbf 0_3 \\
\lfloor{^G\hat{\mathbf p}_{u}\times \rfloor{^{G}_{I_{i,k}}}\hat{\mathbf R}}& \mathbf 0_3& \mathbf 0_3 & {^{G}_{I_{i,k}}\hat{\mathbf R}}
\end{bmatrix}
$}
\end{align}

\bibliographystyle{IEEEtran}
\bibliography{main}

\addtolength{\textheight}{-12cm}   
\end{document}